\documentclass[conference]{IEEEtran}
\IEEEoverridecommandlockouts
\usepackage{cite}
\usepackage{textcomp}
\usepackage{xcolor}

\usepackage{float}
\usepackage{booktabs} 
\usepackage{amsmath,amssymb,amsfonts}
\usepackage{multirow}
\usepackage{graphicx}
\usepackage{algorithm}
\usepackage{algorithmic}
\usepackage{tabularx}
\usepackage{subcaption}
\usepackage{mwe}
\usepackage{times}

\def\BibTeX{{\rm B\kern-.05em{\sc i\kern-.025em b}\kern-.08em
    T\kern-.1667em\lower.7ex\hbox{E}\kern-.125emX}}
\begin{document}

\title{Using Neural Networks and Diversifying Differential Evolution for Dynamic Optimisation}

\author{\IEEEauthorblockN{Maryam Hasani Shoreh}
\IEEEauthorblockA{\textit{Optimisation and Logistics} \\
\textit{University of Adelaide}\\
Adelaide, Australia \\
maryam.hasanishoreh@adelaide.edu.au}
\and
\IEEEauthorblockN{Renato Hermoza Aragonés}
\IEEEauthorblockA{\textit{Australian Institute for Machine Learning} \\
\textit{University of Adelaide}\\
Adelaide, Australia \\
renato.hermozaargones@adelaide.edu.au}
\and
\IEEEauthorblockN{Frank Neumann}
\IEEEauthorblockA{\textit{Optimisation and Logistics} \\
\textit{University of Adelaide}\\
Adelaide, Australia \\
frank.neumann@adelaide.edu.au}
}

\maketitle

\begin{abstract}
Dynamic optimisation occurs in a variety of real-world problems. To tackle these problems, evolutionary algorithms have been extensively used due to their effectiveness and minimum design effort.
However, for dynamic problems, extra mechanisms are required on top of standard evolutionary algorithms.
Among them, diversity mechanisms have proven to be competitive in handling dynamism, and recently, the use of neural networks have become popular for this purpose.
Considering the complexity of using neural networks in the process compared to simple diversity mechanisms, we investigate whether they are competitive and the possibility of integrating them to improve the results. 
However, for a fair comparison, we need to consider the same time budget for each algorithm. Thus instead of the usual number of fitness evaluations as the measure for the available time between changes, we use wall clock timing.
The results show the significance of the improvement when integrating the neural network and diversity mechanisms depends to the type and the frequency of changes.
Moreover, we observe that for differential evolution, having a proper diversity in population when using neural network plays a key role in the neural network's ability to improve the results.
\end{abstract}


\section{Introduction}
\label{sec:intro}
The dynamic environments occur in many real-world problems originating from factors such as variation in the demand market, unpredicted events or variable resources~\cite{branke2003designing,liu2008adaptive}. 
The goal in these dynamic problems is to find the optimum in each instance of the dynamic problem, given a limited computational budget. One approach is to apply an independent optimisation method to separately solve each problem instance. However, a more efficient approach is to solve them through an ongoing search, in which the algorithm detects and responds to changes dynamically~\cite{Nguyen20121}. 
Mathematically, the objective is to find a solution vector ($\vec{x} \in \mathbb{R}^D $) at each time period $t$ such that: $\min_{\vec{x}\in F_t} f(\vec{x}, t)$, where $f:S \rightarrow \mathbb{R}$ is a single objective function, and $t \in N^+$ is the current time period.  
$F_{t}=\{ \vec{x} \mid \vec{x} \in [L,U], g_i (\vec{x},t) \le 0$\}  is the feasible region at time $t$, where $L$ and $U$ are the boundaries of the search space and $g_i(x, t)$ is the linear $i$th inequality constraint.

Standard evolutionary algorithms (EAs) can be easily modified to include change detection and the ability to react to the changes to handle dynamic environments. 
Among the many approaches proposed for reacting to the changes, diversity mechanisms~\cite{Bui2005,Goh_2009} are the simplest and most popular. 
A recent study has revealed how common diversity mechanisms can significantly enhance the performance of a baseline differential evolution (DE) for different environmental changes~\cite{hasani2019use}.
Other approaches include memory-based approaches~\cite{Richter2013}, multi-population approaches~\cite{branke2000multi}, and prediction methods~\cite{Bu_2016}.

Previous work on prediction approaches has shown that they can be well suited to dealing with dynamic problems where there is a trend in the environmental changes~\cite{meier2018prediction}.
For instance, in~\cite{kalman2008tracking}, a Kalman filter is applied to model the movement of the optimum and predict the possible optimum in future environments. Similarly, in~\cite{markov2008evolutionary}, linear regression is used to estimate the time of the next change and Markov chains are adopted to predict new optimum based on the previous time's optimum. Likewise, in~\cite{zhou2013population}, the centre points of Pareto sets in past environments are used as data to simulate the change pattern of those centre points, using a regression model. Besides these methods, neural networks (NNs) have gained increasing attention in recent years~\cite{jiang2017transfer,liu2019neural,meier2018prediction,meier2019uncertaint}.
In~\cite{meier2019uncertaint}, a temporal convolutional network with Monte Carlo dropout is used to predict the next optimum position. The authors propose to control the influence of the prediction through estimation of the prediction uncertainty.
In~\cite{meier2018prediction}, a recurrent NN is proposed that is best suited for objective functions where the optimum movement follows a recurrent pattern.
In other works~\cite{jiang2017transfer,liu2019neural}, where the change pattern is not stable, the authors propose directly constructing a transfer model of the solutions and fitness using NNs, considering the correlation and difference between the two consecutive environments. 

However, despite previous attempts, there are still some concerns regarding the application of NNs to the evolution process. As integrating NNs in EAs is more complicated than using standard diversity mechanisms,
the question arises as to whether they enhance the results to an extent that compensates for their complexity.
In addition, previous work has mainly compared prediction-based methods with a baseline algorithm~\cite{hasanishoreh2020neural} and other prediction-based methods~\cite{hasanishoreh2020neural,meier2018prediction}. 
To the best of our knowledge, only one recent work considers other mechanisms for dynamic handling in comparison with prediction~\cite{meier2019uncertaint}.
However, the time spent by NN has not been accounted for.

We believe that to compare NN to other standard methods fairly, the relative time consumption of NN needs to be accounted for, as this may create a noticeable overhead in the optimisation process; time costs can accrue across the following stages: data collection, training and prediction of new solutions. To account for the timing used by NN, we create a change after an actual running time of the algorithm, instead of the usual number of fitness evaluation as a measure for the available time between changes.
Therefore, to evaluate the effectiveness of NN in the described setting, this work compares common diversity mechanisms using a DE algorithm with and without NN. 
We select DE for our baseline, as it is a competitive algorithm in constrained and dynamic optimisation~\cite{ameca2018comparison}. 
We try to answer the following questions, taking into account the time spent on NN: 
\begin{itemize}
    \item  How does NN compare with other simpler mechanisms for diversifying DE in order to handle dynamic environments?
    \item  Does diversity of population in DE play a role in the effectiveness of NN? 
    \item  Do different frequencies of change impact the suitability of NN in comparison to diversity mechanisms?
\end{itemize}

 
The results of our study show that the extent of the improvement when integrating the neural network and diversity mechanisms depends on the type and the frequency of environmental changes.
In addition, we observe that having a sound diversity in the population has a significant impact on the effectiveness of NN.
The remainder of the paper is as follows. Section~\ref{sec:Prim} introduces preliminaries followed by our experimental methodology in Section~\ref{sec:ExperimentalSetup}. In Section~\ref{sec:crosscompare}, a comparison across all the methods is presented.
In Section~\ref{sec:detailresults}, we carry out detailed investigations based on each diversity variant.
Finally, we finish with some conclusions and elaborate directions for future work.

\section{Preliminaries}
\label{sec:Prim}
In this section, an overview of differential evolution (DE), diversity mechanisms and the neural network (NN) structure are presented.




\subsection{Differential evolution}
\label{subsec:DE}
Differential evolution (DE) is a stochastic search algorithm that is simple, reliable and fast which showed competitive results in constrained and dynamic optimisation~\cite{ameca2018comparison}. Each vector $\vec{x}_{i, G}$ in the current population (called as target vector at the moment of the reproduction) generates one trial vector $\vec{u}_{i, G}$, using a mutant vector $\vec{v}_{i,G}$. The mutant vector is created applying $\vec{v}_{i,G}= \vec{x}_{r0,G} + F (\vec{x}_{r1,G} - \vec{x}_{r2,G})$,
where $\vec{x}_{r0,G}$, $\vec{x}_{r1,G}$, and $\vec{x}_{r2,G}$ are randomly chosen vectors from the current population ($r0 \neq r1 \neq r2 \neq i$); $\vec{x}_{r0,G}$ is known as the base vector and $\vec{x}_{r1,G}$, and $\vec{x}_{r2,G}$ are the difference vectors and $F>0$ is a parameter called scale factor. The trial vector is created by the recombination of the target vector and mutant vector, using a crossover probability $CR \in [0,1]$. 
In this paper, a simple version of DE called DE/rand/1/bin variant is chosen; where ``rand" indicates how the base vector is chosen, ``1" represents  how many vector pairs will contribute in differential mutation, and ``bin" is the crossover type (binomial in our case).
Feasibility rules~\cite{deb2000efficient} is employed for the constraint handling. 
For change detection, we apply the common re-evaluation of the solutions~\cite{Nguyen20121}. In this method, the algorithm regularly re-evaluates specific solutions (in this work, the first and the middle individual of the population) to detect changes in their function values, as well as the constraints.
For change reaction mechanism, we apply two approaches and besides them diversity mechanisms are considered (Section~\ref{sec:div}).
In the first approach, called noNN, the whole population is re-evaluated. In the second approach, called as NN, a number of worst individuals (in terms of objective values and constraint violation based on feasibility rules~\cite{deb2000efficient}) will be replaced with the predicted solutions and the rest of the individuals are re-evaluated.

\subsection{Diversity mechanisms}
\label{sec:div}
We applied the most common diversity mechanisms. For a recent survey regarding the effect of diversity mechanisms in dynamic constrained optimisation see~\cite{hasani2019use}. 

\textbf{Crowding:}
 Among the many niching methods~\footnote{Niching techniques are the extension of standard EAs to multi-modal domains}, we choose the standard crowding method~\cite{sareni1998fitness}. In this method, similar individuals of population are avoided, creating genotypic diversity~\footnote{Diversity can be defined at distinct levels; genotypic level refers to differences among individuals over $\vec{x}$ values}. Instead of competition with the parent, the offspring competes with its closest individual in terms of Euclidean distance. As our problem dimension is high and due to the selected $CR$ for DE, often the parent is the closest individual to the offspring. To avoid selecting the parent, we modified the method such that the offspring competes with N closest individuals (denoted by CwN).
\begin{figure}[t]
\centering
\centerline{\includegraphics[height=1.05in]{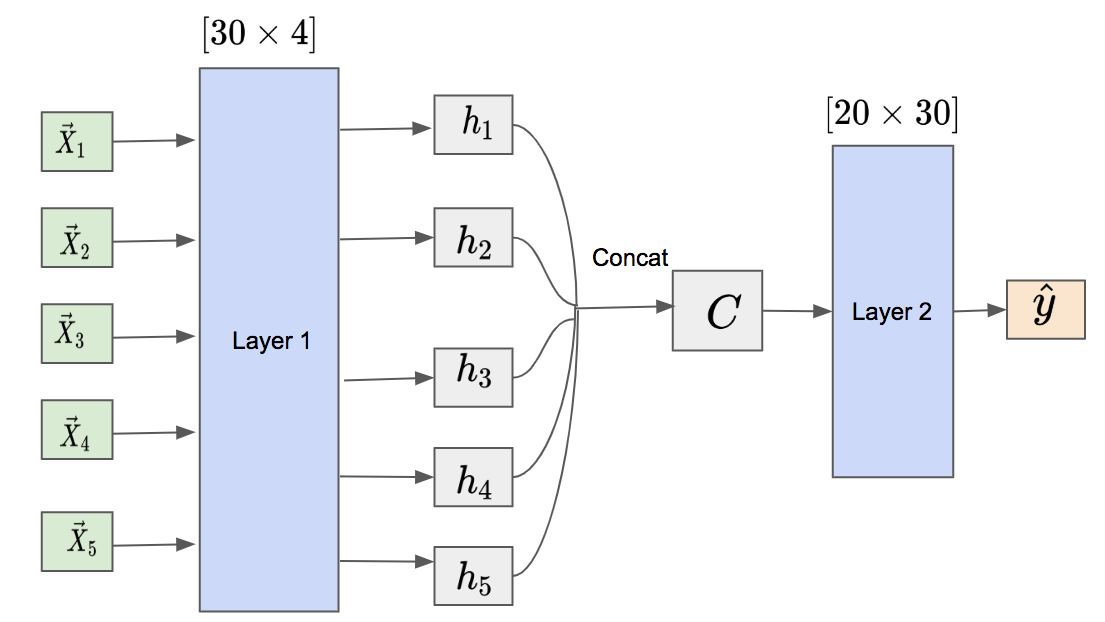}}
    \caption{\scriptsize Structure of neural network}
    \label{fig:NNstructure}
\end{figure}

\textbf{Random immigrants:}
This method (denoted as RI) replaces a certain number of individuals ($replacement\_rate$) with random solutions in the population at each generation to assure continuous exploration~\cite{grefenstette1992genetic}. For our version, random solutions are inserted only when a change is detected. The reason for this modification is that as we consider wall clock timing, if we insert solutions at each generation, there remains insufficient time for the evolution process and the results are adversely affected. 


\textbf{Restart population:}
In this method (denoted by Rst), the population is re-started by random individuals. This is an extreme case of RI by considering the $replacement\_ rate$ to the population size.

\textbf{Hyper-mutation:}
This method was first proposed in genetic algorithms, using an adaptive mutation operator to solve dynamic constrained optimisation problems~\cite{Cobb1990AnII}. Later, it was used for DE in~\cite{ameca2014differential};
after a change detection, the DE parameters (CR and F) change for a number of generations defined empirically (dependent on $\tau$) to favor larger movements.
For our version, denoted by HMu, in addition to changes of the DE parameters, we insert a number of random individuals to the population to assure population diversity.

\subsection{Structure of the neural network}
\label{subsec:NN}
Neural network (NN) is used to predict the future optimum position. To do so, the best solutions of the previous change periods achieved by DE are required to build a time series. Using them NN will go through a training process to learn the change pattern of the optimum position.
Among many structures proposed for NN, we use a simple multi-layer feed forward NN in this work.
We consider a procedure proposed in~\cite{hasanishoreh2020neural} for sample collection. In this work, to expedite sample collection, it is proposed to use $k$-best individuals of each time for a number of previous times ($n_t=5$). Then, a combination of all possibilities ($k^{n_t}$) to build training data is considered. However, the number of samples collected are limited by choosing a random subset of the aforementioned combination. Otherwise, the time spent for training data exponentially increases due to the large number of collected samples.
As for the first environmental changes a small number of samples are existed, hence, it is difficult for the NN to generalize from these data.
To avoid this situation, we have to wait until a minimum amount of samples are collected (defined as $min\_batch$ size).

Figure~\ref{fig:NNstructure} shows the structure of the applied NN that has two hidden layers.
The first layer takes an individual position $\vec{X_i}$ with $d$ dimensions as an input and outputs a hidden representation $h_i$ of the individual with four dimensions.
As the network uses the last five times best individuals to predict a next one, the first layer is applied to each of these five individuals ${\vec{X_1},...,\vec{X_5}}$, independently.
As a result, we obtain 5 hidden representation with four dimensions ${h_1,...,h_5}$; to aggregate their information, we concatenate them into a variable $H$ with $4 \times 5$ dimensions. 
The second layer takes $H$ as input and then outputs a prediction with $d$ dimensions, representing the next best individual.
The first layer employs rectified linear units (ReLU) activation function and the second layer has a linear output without an activation function.
To train the network, we use mean squared error as a loss function. 
A number of neighbouring positions ($n_p$) of the predicted solution (created by adding noise to the original predicted solution) are then replacing the worst individuals of the population in DE to intensify the search in that region of the solution space.

\section{Experimental Methodology}
\label{sec:ExperimentalSetup}
In this section, the test problems, the algorithms' parameter settings, and the performance metrics are summarized.
\subsection{Test problems and parameter settings}
To test our algorithms, we designed the environmental changes in two general cases for common functions: sphere (uni-modal), Rosenbrock (non-separable) and Rastrigin (multi-modal), see details in Table~\ref{tab:test problem design}. 
In the first two experiments, the changes are targeted on $b$ values of one linear constraint in the the form of $\Sigma a_ix_i \le b$~\cite{hasani2019behaviour}. In the last two experiments, the optimum position is transformed based on specific patterns.
\begin{table*}[t]
\centering
\caption{\scriptsize Designed test problems}
\scalebox{0.8}{
\begin{tabular}{l|l}
\textbf{exp1} &  Uniformly random changes on the boundaries of one linear constraint: $b[t+1]=b[t]+\mathcal{U}(lk,uk)$\\\hline 
\textbf{exp2} &  Patterned sinusoidal changes on the boundaries of one linear constraint: $b[t+1]=p\cdot sin(b[t])+\mathcal{N}(0.5)$\\\hline
\textbf{exp3} &  Linear transformation of the optimum position: $\vec{X_{t+1}}=\vec{X_{t}}+0.1t$\\\hline
\textbf{exp4} & Transformation of the optimum position in sinusoidal pattern with random amplitudes: $\vec{X_{t+1}}=\vec{X_{t}}+p[t]sin(\frac{\pi}{2}t)$ 
\end{tabular}
\label{tab:test problem design}
}
\end{table*}
%
\begin{figure*}[t]
\centerline{\includegraphics[width=0.8\textwidth]{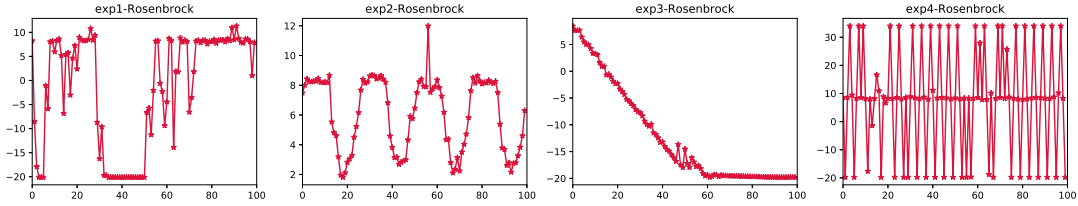}}
    \caption{\scriptsize PCA plot of best\_known positions for each experiment over time}
    \label{fig:pcaPlot}
\end{figure*}
Figure~\ref{fig:pcaPlot} shows the pattern of optimum position~\footnote{Best\_known solutions of each time retrieved by executing baseline DE for 100,000 $runs$.} changes for the Rosenbrock function in each experiment, using principal component analysis (PCA). 

The frequency of change ($\tau$) represents the width for which each time lasts. We indicate that when referring to higher frequencies of change, we are talking about lower values of $\tau$.
Different frequencies of change (1, 5, 10 and 20) will be tested. As we will consider wall clock time, so these values represent time in seconds between the changes. To provide an idea based of number of fitness evaluations for the baseline algorithm: $1 \approx 2000$, $5 \approx 11000$, $10 \approx 22000$, $20 \approx 45000$. Undoubtedly, these numbers are not constant for all test cases, due to the differing time-complexity of each function and the stochastic nature of DE.

The parameters of the problems and algorithms are: $d=30$, $runs=20$ and the number of changes=100.
Parameters of DE are as follows: $NP=20$, $CR=0.3$, $F\sim \mathcal{U}(0.2,0.8)$ and rand/1/bin is the variant of DE~\cite{ameca2018comparison}. The solution space is within $[-5,5]^d$.
For the RI and HMu methods, the $replacement\_rate$ for noNN methods is 7, and for NN methods, it is 2. Since we insert five individuals with NN, we maintain a constant number of individuals in each case overall.
For the HMu method, $F$ and $CR$ are changed for some generations (depending on $\tau$) to $F\sim \mathcal{U}(0.6,0.8)$ and $CR=0.7$, then, after this generation, $6\tau$, they return to normal.

Parameters of NN were selected in a set of preliminary experiments~\cite{hasanishoreh2020neural}: $k=3$, $epochs=4$, $n_w=5$, $batch\_size=4$, $min\_batch=20$ and $n_p=5$.
All the experiments were run on a cluster, allocating one core (2.4GHz) and 4GB of RAM.
Our code is publicly available on GitHub: $https://github.com/renato145/DENN$.
\subsection{Metrics}
\label{subsec:measures}
We applied common metrics in dynamic optimisation as follows:

\textbf{Modified offline error (MOF)}
represents the average of the sum of errors in each generation divided by the total generations~\cite{nguyen2012continuous}.

\begin{equation}
MOF= \frac{1}{G_{max}} \sum_{G = 1}^{G_{max}} (|f(\vec{x}^*,t) - f(\vec{x}_{best,G},t)|)
\label{eq:mof}
\end{equation}
Where $G_{max}$ is the maximum generation, $f(\vec{x}^*,t)$ is the global optimum at current time $t$, and $f(\vec{x}_{best,G},t)$ represents the best solution found so far at generation $G$ at current time $t$.

\textbf{Best error before change (BEBC)} is another common measure that considers the behaviour of algorithm only in the last solution achieved before next change happens. 
\begin{equation}
BEBC= \frac{1}{T_{max}} \sum_{t = 1}^{T_{max}} (|f(\vec{x}^*,t) - f(\vec{x}_{best},t)|)
\label{eq:bebc}
\end{equation}

\textbf{Absolute recovery rate} introduced in~\cite{nguyen2012continuous} is used to analyse the convergence behaviour of the algorithms in dynamic environments. This measure infers how quickly an algorithm starts converging to the optimum before the next change occurs.


%

\begin{equation}
ARR= \frac{1}{T_{max}}\sum_{t = 1}^{T_{max}} (\frac{\sum_{G=1}^{G_{max}(t)}|f_{best}(t,G) - f_{best}(t,1)|}{G_{max}(t)[f^*(t)-f
_{best}(t,1)]})
\label{eq:arr}
\end{equation}

\textbf{Success rate ($SR$)} calculates in how many times (across all times) the algorithm is successful to reach to $\epsilon$-precision from the optimum. 


\begin{table*}[t]
\centering
\caption{\scriptsize Pairwise comparison of methods on MOF values for $\tau=1$ and 20 (mean of 20 $runs$)}
\label{tab:pairwise}
\scalebox{0.7}{
\begin{tabular}{|l|l|ll|ll|ll|ll|ll|}
\hline
Experiment & Function & noNN\_RI & NN\_RI & noNN\_HMu & NN\_HMu & noNN\_No & NN\_No & noNN\_CwN & NN\_CwN & noNN\_Rst & NN\_Rst \\ \hline \hline
\multicolumn{12}{|c|}{$\tau=1$}
\\ \hline
\multirow{3}{*}{1} & Rastrigin & \textbf{100.59} & 103.36 & \textbf{105.7} & 113.28 & 517.9 & \textbf{489.26} & \textbf{222.97} & 376.18 & 144.39 & \textbf{140.48} \\ \cline{2-12} 
 & Rosenbrock & \textbf{73601.01} & 84413.69 & \textbf{64690.02} & 68315.59 & 1958799.29 & \textbf{73601.56} & 624341.74 & \textbf{315775.79} & \textbf{188898.12} & 206130.53 \\ \cline{2-12} 
 & Sphere & \textbf{21.15} & 24.38 & \textbf{20.45} & 21.5 & 546.55 & \textbf{25.67} & 173.99 & \textbf{96.84} & 46.46 & \textbf{41.03} \\ \hline
\multirow{3}{*}{2} & Rastrigin & 84.23 & \textbf{59.59} & 96.73 & \textbf{82.08} & \textbf{30.42} & 30.76 & 196.99 & \textbf{74.28} & 157.8 & \textbf{147} \\ \cline{2-12} 
 & Rosenbrock & 3495.21 & \textbf{2386.98} & 8014.1 & \textbf{3814.8} & 2143.16 & \textbf{619.47} & 73201.31 & \textbf{3838.93} & 17171.11 & \textbf{10804.96} \\ \cline{2-12} 
 & Sphere & 4.25 & \textbf{3.12} & 8.31 & \textbf{4.86} & 9.98 & \textbf{1.29} & 68.38 & \textbf{5.59} & 16.55 & \textbf{10.7} \\ \hline
\multirow{3}{*}{3} & Rastrigin & \textbf{21.06} & 33.74 & \textbf{24.87} & 42.19 & 472.29 & \textbf{318.94} & \textbf{247.64} & 333.28 & \textbf{30.98} & 34.38 \\ \cline{2-12} 
 & Rosenbrock & \textbf{109.23} & 146.53 & \textbf{122.73} & 159.75 & 1787.22 & \textbf{128.55} & 699754.07 & \textbf{1722.27} & 399.02 & \textbf{307.96} \\ \cline{2-12} 
 & Sphere & \textbf{0.09} & 0.11 & \textbf{0.09} & 0.11 & \textbf{0.08} & 0.1 & 133.52 & \textbf{2.03} & 0.36 & \textbf{0.31} \\ \hline
\multirow{3}{*}{4} & Rastrigin & \textbf{841.9} & 871.99 & 772.83 & \textbf{688.72} & 3832 & \textbf{1139.29} & \textbf{1748.45} & 2848.3 & 754.38 & \textbf{716.17} \\ \cline{2-12} 
 & Rosenbrock & 223395465.4 & \textbf{151705461.9} & 189535874 & \textbf{108888429.1} & 957940673.4 & \textbf{205570867.9} & \textbf{477280317} & 945221760 & 175355577.9 & \textbf{127501241} \\ \cline{2-12} 
 & Sphere & 733.49 & \textbf{609.08} & 625.98 & \textbf{428.99} & 3541.08 & \textbf{780.85} & \textbf{1431.8} & 2833.45 & 549.18 & \textbf{492.19} \\ \hline
\multicolumn{12}{|c|}{$\tau=20$}
\\ \hline
\multirow{3}{*}{1} & Rastrigin & 34.84 & \textbf{33.88} & \textbf{40.16} & 46.77 & 516.94 & \textbf{483.62} & \textbf{159.09} & 263.95 & \textbf{46.86} & 47.12 \\ \cline{2-12} 
 & Rosenbrock & \textbf{9210.76} & 11159.41 & 11588.09 & \textbf{10445.52} & 1988246.32 & \textbf{12044.2} & 567717.09 & \textbf{183479.33} & \textbf{19508.38} & 21891.08 \\ \cline{2-12} 
 & Sphere & \textbf{2.86} & 3.44 & 3.56 & \textbf{3.26} & 545.61 & \textbf{3.46} & 143.29 & \textbf{43.39} & \textbf{5.4} & 5.97 \\ \hline
\multirow{3}{*}{2} & Rastrigin & 20.78 & \textbf{19.82} & 25.47 & \textbf{23.21} & 42.58 & \textbf{20.83} & 121.64 & \textbf{21.71} & 46.56 & \textbf{40.38} \\ \cline{2-12} 
 & Rosenbrock & 445.51 & \textbf{290.08} & 1555.23 & \textbf{624.17} & 105.43 & \textbf{88.32} & 24211.28 & \textbf{1307.21} & 2762.26 & \textbf{1549.64} \\ \cline{2-12} 
 & Sphere & 0.51 & \textbf{0.41} & 1.38 & \textbf{0.77} & 12.87 & \textbf{0.19} & 35.24 & \textbf{2.38} & 2.17 & \textbf{1.41} \\ \hline
\multirow{3}{*}{3} & Rastrigin & \textbf{11.54} & 51.33 & \textbf{31.99} & 118.2 & 874.81 & \textbf{802.81} & \textbf{86.23} & 430.54 & 10.41 & \textbf{8.69} \\ \cline{2-12} 
 & Rosenbrock & 22.3 & \textbf{16.06} & 21.73 & \textbf{17.1} & 74.6 & \textbf{17.49} & 204135.11 & \textbf{274.62} & \textbf{11.34} & 16.79 \\ \cline{2-12} 
 & Sphere & 0.04 & \textbf{0.03} & \textbf{0.03} & 0.07 & 0.05 & \textbf{0.03} & 35.14 & \textbf{0.52} & \textbf{0.02} & \textbf{0.02} \\ \hline
\multirow{3}{*}{4} & Rastrigin & 129.34 & \textbf{118.55} & 131.28 & \textbf{107.88} & 3930.78 & \textbf{320.03} & \textbf{717.04} & 1472.18 & 111.38 & \textbf{103.62} \\ \cline{2-12} 
 & Rosenbrock & 23445061.73 & \textbf{16785682.65} & 20298120.95 & \textbf{12073567.99} & 1322886269 & \textbf{34616064.68} & \textbf{152205349.1} & 391706791.9 & 16297615.87 & \textbf{8864684.95} \\ \cline{2-12} 
 & Sphere & 74.36 & \textbf{57.94} & 68.51 & \textbf{56.91} & 4256.18 & \textbf{129.66} & \textbf{588.71} & 1209.84 & 53.91 & \textbf{42.82} \\ \hline
\end{tabular}
}
\end{table*}

\section{Cross Comparison of Approaches}
\label{sec:crosscompare}
Table~\ref{tab:pairwise} shows the MOF values allowing pairwise comparison of diversity variants with and without NN for $\tau=1$ and 20, respectively. In each set of columns, the better performing algorithm is boldfaced. 
We clearly see that for a small $\tau$, methods employing NN are not competitive with their counterparts in each set of diversity variants. 
However, for large $\tau$, NN variants outperform in more test cases. First, as we train the NN with three best solutions of each time, thus when $\tau$ is smaller, the algorithm is not converged, and the best solutions have a greater distance between each other and may not represent the optimum region properly. 
Furthermore, for higher $\tau$ values, the NN time expenditure is negligible compared to the whole evolution process. Last but not least, NN gives direction to the search besides diversity mechanisms that have mostly a random nature. Thus, integration of NN and diversity variant improves the algorithm in comparison to its baseline diversity variant.

The results emphasize that with increasing $\tau$ from 1 to 20, the MOF values decrease. As the algorithms have more timing budget within each change to evolve the solutions and achieve closer values to the optimum. 
However, the performance of NN\_No in exp3 for the Rastrigin function is an anomaly. 
Looking to the plot for optimum position changes in Figure~\ref{fig:pcaPlot}, there is a linearly decreasing trend in the first half of the time scale and a constant optimum position in the second half. 
As NN does not have the correct prediction when the trend changes, then it is not helpful to DE and with the Rastrigin function with multimodal attribute the algorithm with lack of a diversity mechanism has a high chance of getting stuck in a local optimum. This continues for the following changes as NN relies on the previous time's solutions achieved by DE. If the solutions are far from the optimum, the resulting training data will be poor in quality and, therefore, no longer useful to DE. 
This intensifies in higher $\tau$ values as the population becomes more converged. However, algorithms with diversity variants can avoid the local optimum by promoting diversity. 
This shows the importance of diversity variants in the case of wrong predictions. 

\begin{figure}[t]
\centering
  \includegraphics[height=2.2in]{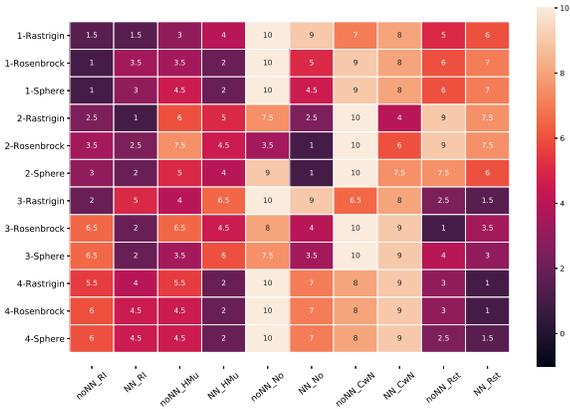} 
  \caption{\scriptsize Heatmap of methods' rank over MOF values ($\tau=5$, and 10). Lower ranks represent better performances. Numbers in Y-axis label show experiment number.}
  \label{fig:heatmap}
\end{figure}

\begin{figure*}[t]
\centerline{\includegraphics[height=1.4in]{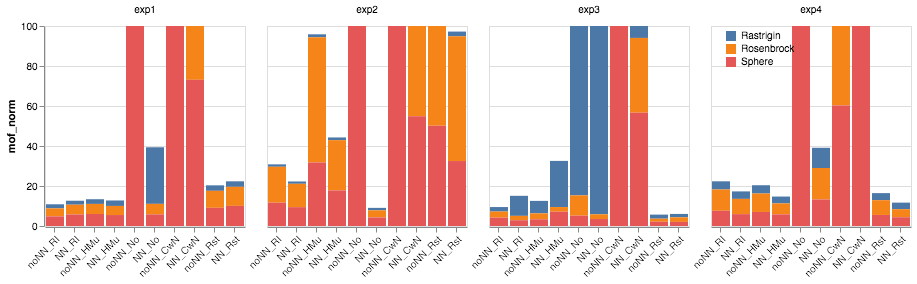}}
    \caption{\scriptsize MOF-norm values for each method and experiment ($\tau=10$), colour-coded with functions}
    \label{fig:chartAll}
\end{figure*}

To compare the algorithms overall, a heatmap with mean rankings of MOF values is presented in Figure~\ref{fig:heatmap}. 
To achieve these scores, we begin by ranking every method, grouping them by function, experiment and frequency. Afterwards, we calculate the mean of the ranks among the frequencies ($\tau=5$ and 10). 
An evident observation is that methods using CwN are not competent in most functions and experiments.
However, this heatmap is not able to define the severity of differences among methods, for relativity analysis we propose Figure~\ref{fig:chartAll}. 
To achieve standard values (denoted as MOF\_norm) in each set of function and experiment, the values are divided by the minimum value among all methods.
To achieve a better resolution in methods' comparison, we limit the y-scale.
In this figure, as an example, we can observe CwN variants and the baseline algorithm (noNN\_No) in all experiments are considerably worse than others, or we can see variants of RI and HMu have better performances overall. 

\begin{figure}[t]
\centering
  \includegraphics[height=2.2in]{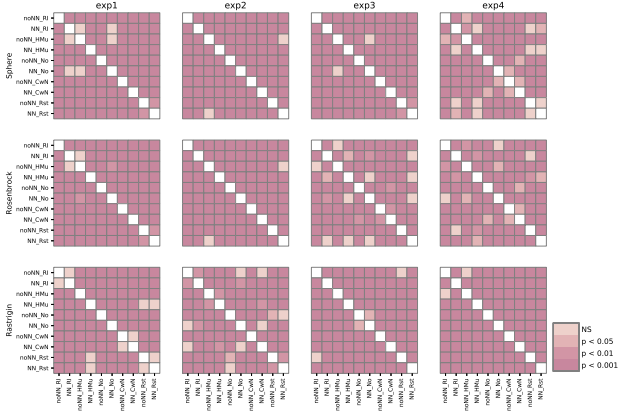} 
  \caption{\scriptsize Kruskal-Wallis statistical test on MOF values ($\tau$=10), NS represents not-significant}
  \label{fig:krus1}
\end{figure}

To validate the results of MOF values, the 95\%-confidence Kruskal-Wallis statistical test and the Bonferroni post hoc test, as suggested in~\cite{Derrac20113} are presented (see Figure~\ref{fig:krus1}). Nonparametric tests were adopted as the samples of runs did not fit a normal distribution based on the Kolmogorov-Smirnov test.
Results of the test display, in most test cases, the methods have significant difference among each
other. 

\begin{figure}[t]
\centerline{\includegraphics[height=1.05in]{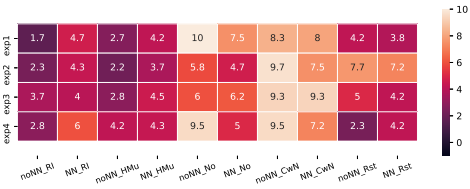}}
    \caption{\scriptsize Heatmap of methods' rank over absolute recovery rate (ARR) values for each experiment ($\tau=5$ and 10). Lower ranks represent better performances}
    \label{fig:arr}
\end{figure}

Due to space limitations, we discard BEBC results, but overall, when $\tau$ is small, this measure shows more differences between methods. However, for $\tau=10$ and 20 most of the methods are able to achieve near to optimum solutions regardless of the differences among their evolution process, an exception is CwN variants with lower performances. In addition, for the Rastrigin with multimodal characteristic the differences between methods for this measure are bigger as there are more chances of algorithms with lack of diversity follow a local optimum and become unable to reach near global optimum solutions.
Figure~\ref{fig:arr} shows a heatmap of the algorithms' ranking based on ARR values considering $\tau=5$ and 10. 
The results show RI and HMu variants have almost the best recovery after a change in all experiments. 
As we can interpret from Equation~\ref{eq:arr}, this measure is slightly biased over the first solution achieved. In consequence, according to this measure, algorithms that start with a very poor solution may achieve a higher ARR values than those starting with a better solution.
So if the first best solution is drastically changed for next generations, this measure reports better results. This is the reason noNN\_RI and noNN\_HMu are the best based on this measure.
Conversely, the worst results is for CwN variants (both NN and noNN) and noNN\_No. Comparing NN\_No and noNN\_No, we can conclude how NN improves recovery capabilities of the algorithm, especially for exp1 and exp4 with drastic changes. 
The heatmap for SR values (see Figure~\ref{fig:sr}) illustrates satisfying results for almost all the methods; meaning they can reach to an $\epsilon$-precision (=10\%) of the optimum for almost all the changes. However, CwN-variants and noNN\_No are the exceptions in which SR values are low. In addition, all the methods show difficulty reaching to optimum in exp2 for the Rastrigin function. Moreover, in other experiments all the methods decrease their performance for this function compared to the other two functions. This is attributed by its multimodal characteristic.

The percentages of the time spent for NN compared to the overall optimisation time, regardless of the experiment and function, is for $\tau=1 \approx 10-11\%$, $\tau=5 \approx 2\%$, $\tau=10 \approx 1\%$ and $\tau=20 \approx 0.5\%$. 
This shows when $\tau$ is higher, it is less expensive to use NN in terms of the computational cost. As NN time remains constant, when $\tau$ is small, the proportion of time for evolution process is lower. 


\section{Detailed Examination of the Use of Neural Networks}
\label{sec:detailresults}
In this section, the methods are compared based on diversity variants. 
\subsection{Crowding}
Looking to the heatmap (see Figure~\ref{fig:heatmap}) based on the colours, it is easily noticeable that the variants of this method perform unsatisfactorily. NN can enhance the results for this method, but still remains inferior compared to other methods.
This method behaves particularly poorly in exp2 and exp3, even compared to noNN\_No. By promoting diversity unnecessarily, CwN adversely affects the convergence of the algorithm to new optimum position, which is not too distant from the previous optimum.  
In addition, from Figure~\ref{fig:arr} we can see this variant delays the recovery of the algorithm (the low rank for ARR values in NN\_CwN and noNN\_CwN indicates the algorithm's inability to recover after a change).

\begin{figure}[t]
\centerline{\includegraphics[height=1.55in]{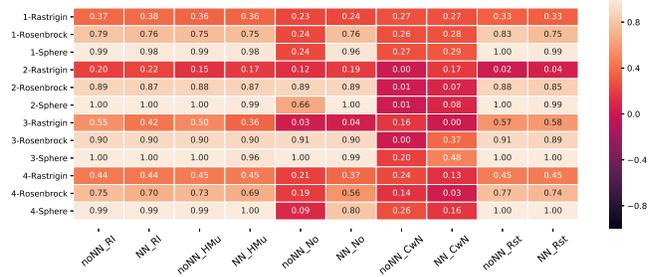}}
    \caption{\scriptsize Heatmap of mean values (20 $runs$) for success rate (SR), considering $\tau=5$ and 10. Higher values represent better performances. Numbers in Y-axis show experiments}
    \label{fig:sr}
\end{figure}

CwN has been reported as one of the best methods for handling dynamic environments in a recent study considering other diversity methods~\cite{hasani2019use}. 
Conversely, CwN is not competitive in our study, because of the following reasons: Firstly, in the evolution process of this method, we need to calculate distances at each generation, and as we are considering wall clock timing, so the algorithm is left with less number of fitness evaluations per each time. 
Secondly, in the literature, this method was most effective for multimodal test problems with several local optima. In such cases, CwN helps to diversify solutions by avoiding similar individuals in each sub-region of the search space~\cite{sareni1998fitness}. In addition, in~\cite{hasani2019use}, CwN demonstrated superior performance for problems that include features such as disconnected and small feasible areas.
Last but not least, in the previous work, CwN was tested on a small problem dimension (2)~\cite{hasani2019use}.
Since our problem's dimension is large, having a $CR$ of 0.3 alter the offspring in only a number of dimensions. In this condition, the closest individual to the offspring often is the parent, causing the method to act in a similar fashion to the no-diversity mechanism, but with an overhead of calculating distances at each iteration. 
To alleviate this, in our CwN version, the offspring will compete with the $N=5$ closest individuals.
Checking the SR-values (see Figure~\ref{fig:sr}) for noNN\_CwN shows this algorithm is barely able to get to the vicinity of optimum ($\epsilon = 10\%$) for almost all the changes in exp2 for all functions and exp3 for the Rosenbrock function. While the other methods are above 90\% for exp2 (the Rosenbrock and sphere).

\subsection{Random immigrants and restart population}
Figure~\ref{fig:heatmap} shows RI variants for most of the experiments and functions rank best amongst all methods. 
The results in Table~\ref{tab:pairwise} show, for large $\tau$, NN outperforms noNN in most of the functions and experiments except for the Rastrigin in exp3, and the Rosenbrock and sphere in exp1 with random changes. The reason is that in noNN\_RI, we only insert random solutions, while for NN\_RI, we diversify population by random immigrants and direct them with the predicted solutions by NN, which expedite the convergence to new optimum leading to better MOF values.
However, for small $\tau$, NN lacks its ability to beat its noNN counterpart. Low timing budget leave the algorithm with poor final solutions at each time. Thus, NN is trained with low quality solutions and is not able to predict the correct future optimum positions.

Although worth to mention, RI is reported to have poor performance in cases of small feasible areas~\cite{hasani2019use}. That is because, the inserted solutions are discarded by constraint handling technique and can not proceed as best solution to guide the search for next generations. In our test problem, we lack such a small feasible area in which RI may show its worst performance. For future work, we will test this method for smaller feasible regions, with features of disconnected feasible areas. 

Comparing NN\_No with NN\_RI to see the effect of RI in comparison with pure NN, we see depending to the type and frequency of changes, their competency differs. For instance, for small $\tau$, for exp1 and exp4 with more random and drastic changes respectively, NN\_RI outperforms. While for exp2 and exp3, with smaller changes, NN\_No wins. However, for large $\tau$, most of the time NN\_RI beats NN\_No.
Looking to Table~\ref{tab:pairwise}, if we compare NN\_RI, NN\_Rst and NN\_No, we can conclude the following.
In most of the experiments, we can clearly observe the behavior of RI is significantly better than Rst method. However, noNN\_Rst is ranked as the best for exp4 with drastic changes. 
Conversely, noNN\_Rst achieves a low ranking for exp2 and exp3, in which the optimum position changes are not huge and by discarding the previous attempts of the algorithm, the performance degrades. 
This implies when using NN, diversity of population is significantly important. In case of NN\_Rst, we have a population scattered around the search space. This is not helpful when NN tries to direct the population toward the new optimum. In case of NN\_RI, however, we have a proper amount of diversity among population (five individuals from NN, and two individuals from RI), so the results are promising.
This explains why noNN\_No has an inferior performance for exp4, in which for larger changes it can not promote diversity to reach optimum. 
While for exp2 and exp3 it is ranked better since it does not need a drastic change in the position. 
\subsection{Hyper-mutation}
HMu's results are quite similar to RI variant with slightly better performances for exp1 and exp4, and worse performances for exp2 and exp3. The rankings in the heatmap~\ref{fig:heatmap} clarify this observation. The reason lies on the shape of the environmental changes for different experiments. Exp1 and exp4 have bigger changes, in which HMu, with larger scale factor after a change, converges faster to the new optimum. While for exp2 and exp3, RI handles the smaller change more efficiently. Since considering hyper mutation factor in HMu (larger $F$ and $CR$), many of the individuals go through a change and convergence is delayed. Whilst, in these two experiments, the optimum position changes are minor. 
%
Due to the same reason, NN\_No also outperforms NN\_HMu for exp2 and exp3. 
This means diversity mechanisms do not always improve algorithm performance when used on top of NN. 

In the proposed version of HMu for DE in~\cite{ameca2014differential}, it is proposed to change DE variant from DE/rand/1/bin to DE/best/1/bin (see Section~\ref{subsec:DE}), when hyper parameters of DE are activated. 
The test problem of that work is two-dimensional. For larger problem sizes (such as our work), this method is not able to promote diversity. This is because in a DE promoting diversity needs a minimum level of population diversity. Thus, in a set of preliminary experiments, we select to insert random solutions (seven individuals for the case without NN, and two for the case using NN). 
Based on the statistical test results, this method does not have significant difference in some cases with RI (for exp1 sphere and rosenbrock), and in some cases with Rst (exp1 the Rastrigin and exp2 sphere).

\section{Conclusions and future directions}
\label{sec:conc}
In this work, a NN was considered to solve dynamic problems together with DE.
Given the complexity of integrating NN into the evolution process and considering the time spent to train it, we investigated weather it can be competitive compared to standard diversity mechanisms.
We empirically studied the possibility of integrating them to extract the best of each to improve the results. 
We observed that diversity of population is essential when using NN for DE. 
Since evolution process of DE algorithm depends on the diversity of population, if we use NN without other diversity mechanism, as the inserted solutions are distributed around the predicted value, then it is slower to explore the search space leading to lower MOF values.
In addition, in some cases due to the multi-modality of the function, the algorithm may get stuck in a local optimum and produce poor samples for NN to train. Hence, it is important to have a diversity mechanism. 
On the other hand, we observed that NN can improve the results of simple diversity mechanisms by directing the search toward the next optimum besides random nature of diversity mechanisms.
The presented results was for one simple feed-forward NN, however, considering several proposed structures of NNs in literature, their application to handle dynamic optimisation problems is in its infancy yet. So for future work, we encourage application of other NN designs to investigate their differences based of a range of problem features.

\section*{Acknowledgment}
This work has been supported through Australian Research Council (ARC) grants DP160102401 and DP180103232.

%

\bibliographystyle{IEEEtran}
\bibliography{bibliography}

\begin{thebibliography}{10}
\providecommand{\url}[1]{#1}
\csname url@samestyle\endcsname
\providecommand{\newblock}{\relax}
\providecommand{\bibinfo}[2]{#2}
\providecommand{\BIBentrySTDinterwordspacing}{\spaceskip=0pt\relax}
\providecommand{\BIBentryALTinterwordstretchfactor}{4}
\providecommand{\BIBentryALTinterwordspacing}{\spaceskip=\fontdimen2\font plus
\BIBentryALTinterwordstretchfactor\fontdimen3\font minus
  \fontdimen4\font\relax}
\providecommand{\BIBforeignlanguage}[2]{{%
\expandafter\ifx\csname l@#1\endcsname\relax
\typeout{** WARNING: IEEEtran.bst: No hyphenation pattern has been}%
\typeout{** loaded for the language `#1'. Using the pattern for}%
\typeout{** the default language instead.}%
\else
\language=\csname l@#1\endcsname
\fi
#2}}
\providecommand{\BIBdecl}{\relax}
\BIBdecl

\bibitem{branke2003designing}
J.~Branke and H.~Schmeck, ``Designing evolutionary algorithms for dynamic
  optimization problems,'' in \emph{Advances in evolutionary computing}.\hskip
  1em plus 0.5em minus 0.4em\relax Springer, 2003, pp. 239--262.

\bibitem{liu2008adaptive}
L.~Liu, E.~M. Zechman, E.~D. Brill, Jr, G.~Mahinthakumar, S.~Ranjithan, and
  J.~Uber, ``Adaptive contamination source identification in water distribution
  systems using an evolutionary algorithm-based dynamic optimization
  procedure,'' in \emph{Water Distribution Systems Analysis Symposium 2006},
  2008, pp. 1--9.

\bibitem{Nguyen20121}
T.~Nguyen, S.~Yang, and J.~Branke, ``Evolutionary dynamic optimization: A
  survey of the state of the art,'' \emph{Swarm and Evolutionary Computation},
  vol.~6, no.~0, pp. 1 -- 24, 2012.

\bibitem{Bui2005}
L.~T. Bui, H.~A. Abbass, and J.~Branke, ``Multiobjective optimization for
  dynamic environments,'' in \emph{2005 IEEE Congress on Evolutionary
  Computation}, vol.~3, Sept 2005, pp. 2349--2356 Vol. 3.

\bibitem{Goh_2009}
C.~K. Goh and K.~C. Tan, ``A competitive-cooperative coevolutionary paradigm
  for dynamic multiobjective optimization,'' \emph{IEEE Transactions on
  Evolutionary Computation}, vol.~13, no.~1, pp. 103--127, Feb 2009.

\bibitem{hasani2019use}
M.~Hasani-Shoreh and F.~Neumann, ``On the use of diversity mechanisms in
  dynamic constrained continuous optimization,'' in \emph{International
  Conference on Neural Information Processing}.\hskip 1em plus 0.5em minus
  0.4em\relax Springer, 2019, pp. 644--657.

\bibitem{Richter2013}
H.~Richter, \emph{Evolutionary Computation for Dynamic Optimization
  Problems}.\hskip 1em plus 0.5em minus 0.4em\relax Berlin, Heidelberg:
  Springer Berlin Heidelberg, 2013, ch. Dynamic Fitness Landscape Analysis, pp.
  269--297.

\bibitem{branke2000multi}
J.~Branke, T.~Kau{\ss}ler, C.~Smidt, and H.~Schmeck, ``A multi-population
  approach to dynamic optimization problems,'' in \emph{Evolutionary design and
  manufacture}.\hskip 1em plus 0.5em minus 0.4em\relax Springer, 2000, pp.
  299--307.

\bibitem{Bu_2016}
C.~Bu, W.~Luo, and L.~Yue, ``Continuous dynamic constrained optimization with
  ensemble of locating and tracking feasible regions strategies,'' \emph{IEEE
  Transactions on Evolutionary Computation}, vol.~PP, no.~99, pp. 1--1, 2016.

\bibitem{meier2018prediction}
A.~Meier and O.~Kramer, ``Prediction with recurrent neural networks in
  evolutionary dynamic optimization,'' in \emph{International Conference on the
  Applications of Evolutionary Computation}.\hskip 1em plus 0.5em minus
  0.4em\relax Springer, 2018, pp. 848--863.

\bibitem{kalman2008tracking}
C.~Rossi, M.~Abderrahim, and J.~C. D{\'\i}az, ``Tracking moving optima using
  kalman-based predictions,'' \emph{Evolutionary computation}, vol.~16, no.~1,
  pp. 1--30, 2008.

\bibitem{markov2008evolutionary}
A.~Sim{\~o}es and E.~Costa, ``Evolutionary algorithms for dynamic environments:
  prediction using linear regression and markov chains,'' in
  \emph{International Conference on Parallel Problem Solving from
  Nature}.\hskip 1em plus 0.5em minus 0.4em\relax Springer, 2008, pp. 306--315.

\bibitem{zhou2013population}
A.~Zhou, Y.~Jin, and Q.~Zhang, ``A population prediction strategy for
  evolutionary dynamic multiobjective optimization,'' \emph{IEEE transactions
  on cybernetics}, vol.~44, no.~1, pp. 40--53, 2013.

\bibitem{jiang2017transfer}
M.~Jiang, Z.~Huang, L.~Qiu, W.~Huang, and G.~G. Yen, ``Transfer learning-based
  dynamic multiobjective optimization algorithms,'' \emph{IEEE Transactions on
  Evolutionary Computation}, vol.~22, no.~4, pp. 501--514, 2017.

\bibitem{liu2019neural}
X.-F. Liu, Z.-H. Zhan, T.-L. Gu, S.~Kwong, Z.~Lu, H.~B.-L. Duh, and J.~Zhang,
  ``Neural network-based information transfer for dynamic optimization,''
  \emph{IEEE transactions on neural networks and learning systems}, 2019.

\bibitem{meier2019uncertaint}
A.~Meier and O.~Kramer, ``Predictive uncertainty estimation with temporal
  convolutional networks for dynamic evolutionary optimization,'' in
  \emph{International Conference on Artificial Neural Networks}.\hskip 1em plus
  0.5em minus 0.4em\relax Springer, 2019, pp. 409--421.

\bibitem{hasanishoreh2020neural}
M.~Hasani-Shoreh, R.~H. Aragonés, and F.~Neumann, ``Neural networks in
  evolutionary dynamic constrained optimization: Computational cost and
  benefits,'' 2020.

\bibitem{ameca2018comparison}
M.-Y. Ameca-Alducin, M.~Hasani-Shoreh, W.~Blaikie, F.~Neumann, and
  E.~Mezura-Montes, ``A comparison of constraint handling techniques for
  dynamic constrained optimization problems,'' in \emph{2018 IEEE Congress on
  Evolutionary Computation (CEC)}.\hskip 1em plus 0.5em minus 0.4em\relax IEEE,
  2018, pp. 1--8.

\bibitem{deb2000efficient}
K.~Deb, ``An efficient constraint handling method for genetic algorithms,''
  \emph{Computer methods in applied mechanics and engineering}, vol. 186,
  no.~2, pp. 311--338, 2000.

\bibitem{sareni1998fitness}
B.~Sareni and L.~Krahenbuhl, ``Fitness sharing and niching methods revisited,''
  \emph{IEEE transactions on Evolutionary Computation}, vol.~2, no.~3, pp.
  97--106, 1998.

\bibitem{grefenstette1992genetic}
J.~J. Grefenstette \emph{et~al.}, ``Genetic algorithms for changing
  environments,'' in \emph{PPSN}, vol.~2, 1992, pp. 137--144.

\bibitem{Cobb1990AnII}
H.~G. Cobb, ``An investigation into the use of hypermutation as an adaptive
  operator in genetic algorithms having continuous, time-dependent
  nonstationary environments,'' 1990.

\bibitem{ameca2014differential}
M.-Y. Ameca-Alducin, E.~Mezura-Montes, and N.~Cruz-Ramirez, ``Differential
  evolution with combined variants for dynamic constrained optimization,'' in
  \emph{Evolutionary computation (CEC), 2014 IEEE congress on}.\hskip 1em plus
  0.5em minus 0.4em\relax IEEE, 2014, pp. 975--982.

\bibitem{hasani2019behaviour}
M.~Hasani-Shoreh, M.-Y. Ameca-Alducin, W.~Blaikie, and M.~Schoenauer, ``On the
  behaviour of differential evolution for problems with dynamic linear
  constraints,'' in \emph{2019 IEEE Congress on Evolutionary Computation
  (CEC)}.\hskip 1em plus 0.5em minus 0.4em\relax IEEE, 2019, pp. 3045--3052.

\bibitem{nguyen2012continuous}
T.~T. Nguyen and X.~Yao, ``Continuous dynamic constrained optimization—the
  challenges,'' \emph{IEEE Transactions on Evolutionary Computation}, vol.~16,
  no.~6, pp. 769--786, 2012.

\bibitem{Derrac20113}
J.~Derrac, S.~Garc\'ia, D.~Molina, and F.~Herrera, ``A practical tutorial on
  the use of nonparametric statistical tests as a methodology for comparing
  evolutionary and swarm intelligence algorithms,'' \emph{Swarm and
  Evolutionary Computation}, vol.~1, no.~1, pp. 3--18, 2011.

\end{thebibliography}

\end{document}